\pdfoutput=1

\documentclass[11pt]{article}

\usepackage{ACL2023}

\usepackage{times}
\usepackage{latexsym}

\usepackage[T1]{fontenc}

\usepackage[utf8]{inputenc}

\usepackage{microtype}

\usepackage{inconsolata}

\usepackage{bbold}
\usepackage{xcolor}
\usepackage{amsmath}
\usepackage{multirow}
\usepackage[verbose]{placeins}
\usepackage{graphicx}

\title{SongRewriter: A Chinese Song Rewriting System with Controllable Content and Rhyme Scheme}

\author{Yusen Sun\textsuperscript{\rm 1,2,*}, Liangyou Li\textsuperscript{\rm 2}, Qun Liu\textsuperscript{\rm 2}, Dit-Yan Yeung\textsuperscript{\rm 1}\\
        \textsuperscript{\rm 1}The Hong Kong University of Science and Technology, Hong Kong SAR, China  \\
        \textsuperscript{\rm 2}Huawei Noah’s Ark Lab \\
        \texttt{ysunbc@connect.ust.hk}, \texttt{\{liliangyou, liuqun\}@huawei.com}, \texttt{dyyeung@cse.ust.hk} \\
        }

\begin{document}
\maketitle
\begingroup\def\thefootnote{*}\footnotetext{Work done during internship at Huawei Noah’s Ark Lab.}\endgroup

\begin{abstract}
Although lyrics generation has achieved significant progress in recent years, it has limited practical applications because the generated lyrics cannot be performed without composing compatible melodies. In this work, we bridge this practical gap by proposing a song rewriting system which rewrites the lyrics of an existing song such that the generated lyrics are compatible with the rhythm of the existing melody and thus singable. In particular, we propose \textit{SongRewriter},\footnote{ Source code implemented in \href{https://www.mindspore.cn/en}{MindSpore Lite tool} is available at \url{https://github.com/huawei-noah/noah-research/tree/master/NLP/SongRewriter}.} 
a controllable Chinese lyrics generation and editing system which assists users without prior knowledge of melody composition. The system is trained by a randomized multi-level masking strategy which produces a unified model for generating entirely new lyrics or editing a few fragments. To improve the controllabiliy of the generation process, we further incorporate a keyword prompt to control the lexical choices of the content and propose novel decoding constraints and a vowel modeling task to enable flexible end and internal rhyme schemes. While prior rhyming metrics are mainly for rap lyrics, we propose three novel rhyming evaluation metrics for song lyrics. Both automatic and human evaluations show that the proposed model performs better than the state-of-the-art models in both contents and rhyming quality.
\end{abstract}

\begin{figure}[!ht]
\begin{center}
\centerline{\includegraphics[width=\columnwidth]{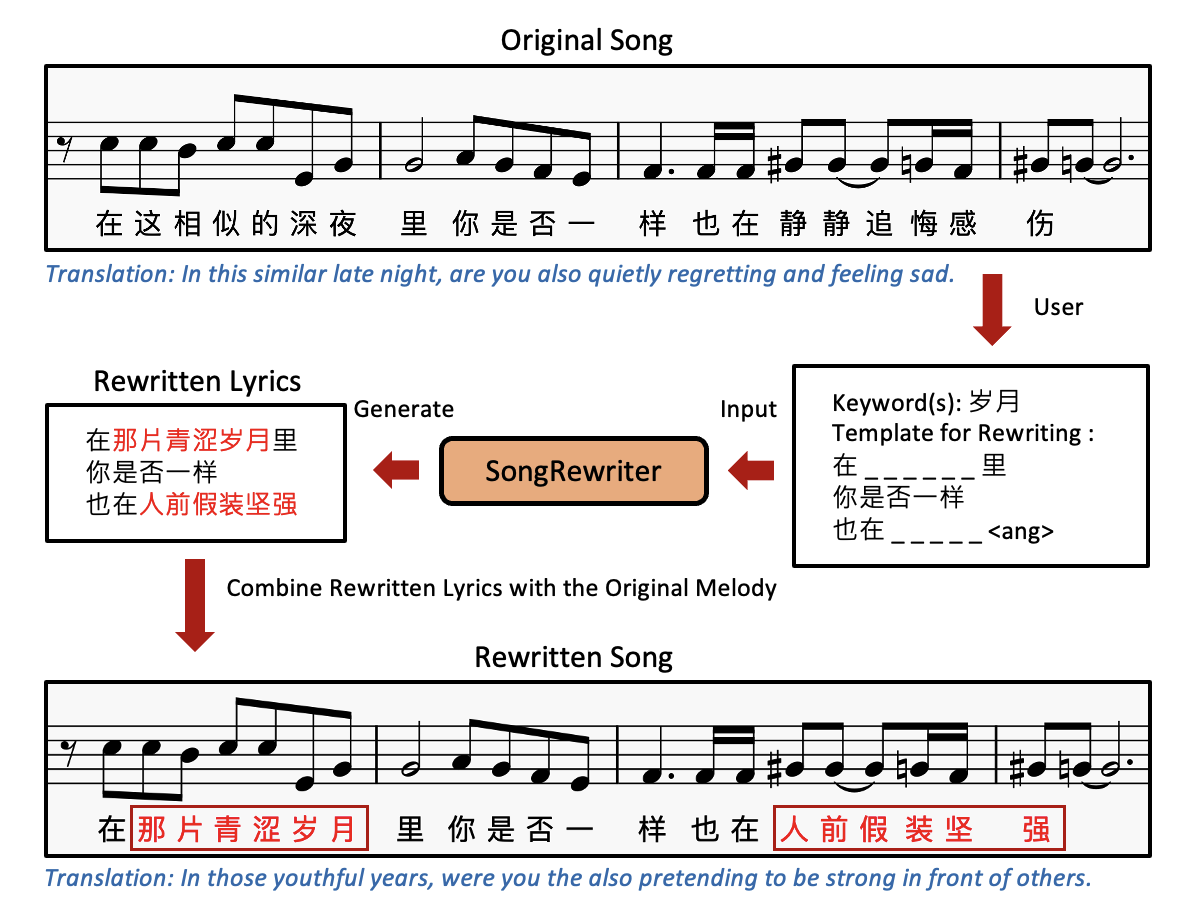}}
\caption{An overview of the proposed Chinese song rewriting system. Given an existing song, the users first mask the part(s) of the lyrics that they want to rewrite. Then, \textit{SongRewriter} generates new lyrics corresponding to the masked fragments. Last, the rewritten lyrics are combined with the original melody to form a new song. During the generation, user can require the content to include specific keywords or control the rhyme scheme by setting the vowels of the output characters at specific positions. }
\label{system-flow}
\end{center}
\end{figure}

\section{Introduction}
With the rapid development in natural language processing \citep{seq2seq, seq2seq_attn, transformer, devlin-etal-2019-bert}, lyrics generation has achieved significant advancement in recent years \citep{potash-etal-2015-ghostwriter, nikolov-etal-2020-rapformer, ai-lyricist, xue-etal-2021-deeprapper}. Prior works mainly focus on two research topics: singability of the generated lyrics \citep{watanabe-etal-2018-melody, lee-etal-2019-icomposer} and controllability of the content and/or rhyme \citep{nikolov-etal-2020-rapformer, xue-etal-2021-deeprapper}.

Current methods of generating singable lyrics is typically conditioned on a given melody. Accordingly, they treat the generation as a sequence-to-sequence translation task. However, there are two main challenges: 1) the parallel lyrics-melody dataset for training the model is limited, which mainly consists of 7,998 songs proposed by  \citet{Lyrics-Conditioned-Neural-Melody-Generation}; 2) melody notes and lyric syllables are loosely correlated and thus the alignments between them are hard to learn from the limited data. Therefore, prior works simplify the task by assuming a one-to-one mapping between the melody notes and lyric syllables \citep{Lyrics-Conditioned-Neural-Melody-Generation, ai-lyricist}. However, such an assumption and restrictions may lead to a sub-optimal performance, as the mapping relationship is usually many-to-one in real life.

Another important research question on lyrics generation is how to control the generated content and rhyme. In prior works, the controllable content is usually achieved by conditioning the generation on given hint words or sentences \citep{shen-etal-2019-controlling, zhang-etal-2022-qiuniu}. However, they usually ignore the requirements of generating lyrics from a draft where the model needs to edit some sentences or words. In addition, prior works on rhyme control mainly focus on rhymes at the end of sentences (\textit{end rhyme}) \citep{potash-etal-2015-ghostwriter, nikolov-etal-2020-rapformer, xue-etal-2021-deeprapper, liu-etal-2022-chipsong}. To the best of our knowledge, no work has been done on both {\em internal rhyme} and {\em end rhyme} schemes.

In this work, we develop a user-assist AI system \textit{SongRewriter} which can generate singable lyrics by rewriting parts of or the whole lyrics of a given song (i.e., {\em partial rewriting} and {\em full rewriting}, respectively) with controllable contents and rhyme schemes, as illustrated in Figure \ref{system-flow}. To address the difficulty of learning the correlation between the melody and lyrics from a limited amount of parallel dataset, we propose to generate lyrics which have the same number of syllables as the original lyrics of the song. Therefore, the generated lyrics can well align with the melody. In addition, this method can directly learn the generation from text data so as to bypass the demand of parallel datasets.

Specifically, we adopt a transformer-based sequence-to-sequence auto-regressive model \citep{transformer} as our model backbone. The model is trained by masking a few random fragments from the encoder's inputs and predicting the masked fragments by the decoder. The generation process can be controlled in terms of three aspects. 1) To enable rewriting arbitrary parts of the lyrics, we train the model by randomly performing one of the three masking (i.e., token-level, sentence-level and song-level) strategies, corresponding to different levels of rewriting tasks.  2) To control the lexical choice of contents, we allow the generation to condition on given keyword prompts. This is achieved by training on extracted keywords from masked positions as additional encoder inputs.  3) To enable lyrics generation with arbitrary pre-defined rhyme schemes with vowel specified, we introduce additional vowel inputs and apply a vowel mask strategy during training. This equips the model with the ability of generating tokens with required vowels at arbitrary positions. Since end rhyme is the most frequently used rhyme type, we specifically incorporate reverse language modelling and propose decoding constraints during inference to improve the vowel consistency and rhyming word diversity.

We evaluate our model on both generation controllability and quality in terms of keyword recall, vowel accuracy, lexical diversity, coherence, perplexity and rhyme quality. Since prior evaluation metrics on rhyme quality are mainly for rap lyrics and ignore the problem of identical rhyming words, we propose three new rhyme metrics to measure the local rhyme between adjacent sentences, global rhyme and diversity of rhyming words, respectively.  Experimental results show that our model outperforms baseline models and state-of-the-art models on both full and partial rewriting tasks.

Our contributions are summarized as follows:
\begin{itemize}
    \item We propose \textit{SongRewriter} which generates melody-aligned lyrics by rewriting the lyrics of songs. It bypasses the difficulties of modelling the melody-lyrics correlation from limited parallel datasets. 
    \item We propose a multi-level randomized masking scheme for training \textit{SongRewriter}, which allows the model to rewrite arbitrary parts of the inputs according to the bidirectional context. 
    \item We introduce a partial vowel masking strategy into training to enable lyrics generation on any rhyme schemes, and we design a novel decoding strategy to improve the end rhyme consistency and rhyming word diversity. 
    \item We propose novel metrics for rhyme evaluation. We collect data from the internet for training and testing. Experiments show the effectiveness of our proposed model. 
\end{itemize}

\begin{figure*}[ht]
\begin{center}
\centerline{\includegraphics[width=2\columnwidth]{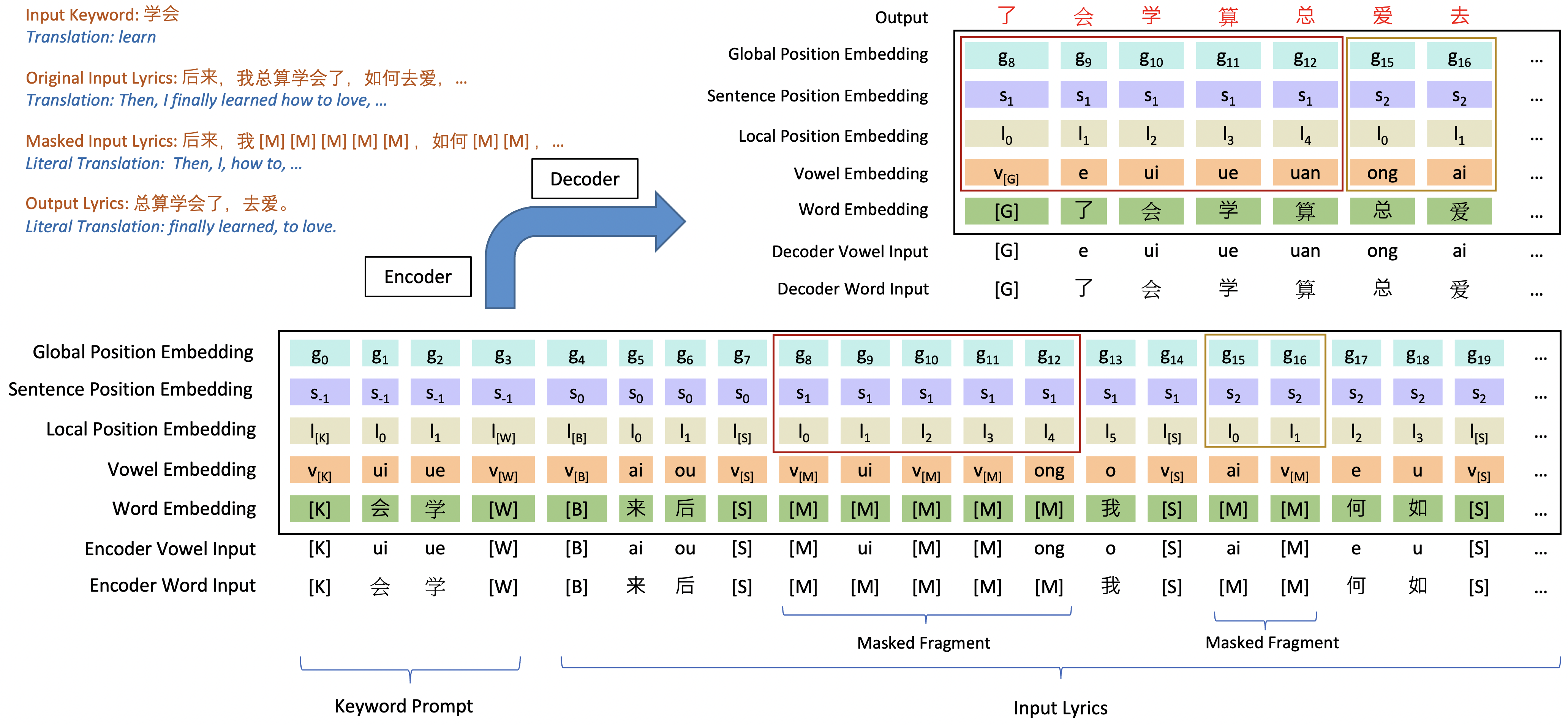}}
\caption{The architecture of the proposed \textit{SongRewriter} model. The inputs of the encoder consist of a keyword prompt and partially masked lyrics. The keywords in the prompt are extracted from the masked fragments during training. The decoder uses [G] as a start token and generates the masked tokens autoregressively. The text order is reversed for end rhyme modeling.   The original lyrics is from \textit{Later} by \textit{Rene Liu}.}
\label{model}
\end{center}
\end{figure*}

\section{Related Work}
Lyrics generation can be mainly divided into two categories, rap lyrics generation
\citep{potash-etal-2015-ghostwriter, dopeLearning, nikolov-etal-2020-rapformer, xue-etal-2021-deeprapper} 
and song lyrics generation
\citep{ramakrishnan-a-etal-2009-automatic, watanabe-etal-2018-melody, Fan2019AHA, lee-etal-2019-icomposer, songmass, ai-lyricist, liu-etal-2022-chipsong}. 
While rap lyrics generation mainly focuses on improving the rhyming performance
\citep{nikolov-etal-2020-rapformer, xue-etal-2021-deeprapper}, 
song lyrics generation concentrates on the alignment between melody and lyrics 
\citep{watanabe-etal-2018-melody, lee-etal-2019-icomposer, songmass}.

Rhyming is an essential element for lyrics and poetry. To model rhyme, current approaches can be mainly divided into three types. 
The first type is to encourage the model to learn the rhyme structure implicitly during the training by adding additional rhyme signals. 
\citet{potash-etal-2015-ghostwriter} append $\langle endLine \rangle$ token to each sentence. 
\citet{li-etal-2020-rigid} employ an additional format embedding as inputs to emphasize the rhyming tokens. 
\citet{lau-etal-2018-deep} incorporate an extra model to encode the ending tokens of the sentences. 
While \citet{zhang-etal-2020-youling} and \citet{xue-etal-2021-deeprapper} both generate the rhyming word before the rest of the sentence, 
\citet{zhang-etal-2020-youling} move the last word of the sentence to the front, and 
\citet{xue-etal-2021-deeprapper} generate sentence from right to left by reversing the word order.
The second type is to apply explicit rhyme constraints during the training to force the model to generate rhyming sentences. 
\citet{jhamtani-etal-2019-learning} apply a discriminator on the sentence-ending words to learn the rhyme pattern adversarially. 
The last type is through post-editing, where the model first generates the lyrics, then another model edits the ending words to fulfil the rhyming constraints \citep{nikolov-etal-2020-rapformer}.  

Most prior works focus on the rhyme at the end of the sentences \citep{xue-etal-2021-deeprapper, liu-etal-2022-chipsong}. In this work, we extend the rhyme control to arbitrary rhyme schemes for both internal rhyme and end rhyme. To the best of our knowledge, this is the first work enabling arbitrary rhyme schemes.  

\section{Method}
The proposed model is a transformer-based auto-regressive sequence-to-sequence model \citep{transformer}. As shown in Figure \ref{model}, given an input of masked lyrics and keyword prompt to the encoder, the decoder generates output tokens corresponding to the masked tokens of the input, which contain the keywords in the prompt and satisfy the vowel constraints. Such generation also improves decoding efficiency and forces the model to rely more on the source input.

In a basic setting, given the lyrics of a song, the tokens of the input lyrics are of the following form, 
$$[B], x_{00}, ..., [M],..., [S],..., x_{ij}, ..., [S], [E]$$
where $x_{ij}$ denotes the $j$th tokens in the $i$th sentence of the lyrics, $[S]$ is the inter-sentence delimiter,  $[B]$ is placed at the beginning of the lyrics with  $[E]$ at the end, and the tokens to be rewritten are replaced by $[M]$ which will be predicted by the decoder.

To enable the controllability of the generation process, we further incorporate the keyword prompt into the model to control the lexical choices, propose the multi-level masking strategy to enable rewriting arbitrary parts of the input in a single model and the rhyme control strategy to inject pre-defined rhyme schemes and improve rhyming word diversity.

\subsection{Keyword Prompt}
Using keyword prompts to control the text generation has been explored in other tasks, such as poetry generation \citep{zhipeng-etal-2019-jiuge}. In the task of lyrics generation, prior works mainly focus on theme control \citep{shen-etal-2019-controlling}.
In this work we introduce the technique of keyword prompts into lyrics generation to force the model generating lyrics containing these keywords.

Specifically, a keyword prompt is prepended to the input lyrics. The keyword prompt is a concatenation of a set of keywords in the following format, 
$$[K], k_{00},..., [W], k_{10}, ..., [W], ..., k_{ij}, ...,   [W]$$ 
where $ k_{ij}$ is the $j$th token in the $i$th keyword , $[W]$ is the inter-keyword delimiter, and  $[K]$ is the start token of the prompt. 

During training, we first  use \textit{jieba}\footnote{\url{https://github.com/fxsjy/jieba}} to segment the  masked fragments into words and obtain their Parts-of-Speech tags. Then, we use the nouns and verbs to form a keyword database. Last, we sample a random number of keywords (ranging from 0 to 5) from the keyword database to form a keyword prompt. During inference, the keyword prompt is optional and can be provided by users.

\subsection{Randomized Multi-Level Masking Scheme}

As song rewriting requires the number of syllables between the original lyrics and the generated lyrics to be identical, we adopt the framework of MASS \citep{mass}, a sequence-to-sequence model pre-training method, which uses the decoder to predict the masked tokens in the encoder. However, MASS trains the model on inputs of single-sentence examples by masking a fragment of continuous tokens (around 50\% of the input). Therefore, it is not an optimal strategy for full lyrics generation and editing arbitrary parts of the input lyrics. 

Accordingly, we propose a novel masking strategy which masks the input from three levels (token-level, sentence-level and song-level) to simulate the partial rewriting and full rewriting tasks. During training, for each input lyric, we randomly apply one of the following masking strategies:
\begin{itemize}
\item \textbf{Token-Level Masking}: To simulate the task of rewriting phrases in a sentence, for each sentence in the input lyric, we mask a few fragments of the sentence with a random ratio and train the model to reconstruct the masked portions of the sentences.
\item \textbf{Sentence-Level Masking}: To enable sentence rewriting, we mask a random ratio of sentences (entire sentences) from the input and train the model to reconstruct the masked sentences.
\item \textbf{Song-Level Masking}: We masks all the input tokens to simulate the full rewriting task.
\end{itemize}

We denote the above three masking schemes as \{TOKEN, SENT, ALL\}, respectively. During training, for SENT,  we sample a masking ratio from a uniform distribution $\operatorname{U}(0, 1)$ and randomly select the corresponding ratio of sentences. For TOKEN, we sample a masking ratio from a uniform distribution $\operatorname{U}(0, 1)$ for each sentence and then randomly select the ratio of tokens to mask.

\subsection{Rhyme Modeling and Control}
Our method for rhyme modelling and control is divided into two parts, rhyming modelling and control for the final syllables of the lines (\textit{end rhyme}), and rhyme control for an arbitrary rhyme scheme which defines the required vowels at specific positions in the generated lyrics (\textit{internal rhyme}).

\subsubsection*{End Rhyme Modeling and Constraint}
\textit{End rhyme} is the most frequently used rhyme type for lyrics and poems. It occurs when the last words of the sentences rhyme. Inspired by the rhyme modelling for rap lyrics \citep{xue-etal-2021-deeprapper}, we adopt reverse language modelling with two additional position embeddings, sentence position embedding and local position embedding, to facilitate the modelling of rhyme features in the lyrics. Specifically, we reverse the order of the characters in each sentence for both inputs and target outputs (while keeping the sentence order unchanged). Therefore, the reverse sentence starts with the potential rhyming character, i.e., the end character in the original sentence. Accordingly, the model can easily learn to identify the rhyming characters from the inputs with the local position l\textsubscript{0}.

However, since rhyming with identical words is considered inferior, we incorporate two control factors at inference time to encourage rhyme consistency and rhyming word diversity. Given the end character set $e_{<t}=\{e_0,...,e_{t-1}\}$ and their corresponding vowel set $v_{<t}=\{v_0,...,v_{t-1}\}$ from the previous $t$ sentences, we define an adjusted probability of the end character of the $(t+1)$th sentence being $x_i$ as, 

\begin{equation}
    \bar{p}^i_t = \frac{p^i_t \cdot  \Lambda(v_{x_i}, v_{<t}) \cdot \Gamma(x_i, e_{<t})}{\sum_j^N p^j_t \cdot  \Lambda(v_{x_j}, v_{<t}) \cdot \Gamma(x_j, e_{<t})}
\end{equation}

\noindent where $p^i_t$ is the predicted probability of the token $x_i$ in the vocabulary, and $v_{x_i}$ is the vowel of the token $x_i$. The two factors:
\begin{itemize}
    \item $\Lambda(\cdot)$ returns $\lambda$ if the vowel of the token $x_i$ appears in the end vowel set $v_{<t}$ of previous sentences; otherwise, returns $1$. 
    \item $\Gamma(\cdot)$ returns $\gamma$ if $x_i$  appears in the  end character set $e_{<t}$ of previous sentences, otherwise $1$.
\end{itemize}
$\lambda$ and $\gamma$ are the hyper-parameters to control the rhyming effect. While a larger $\lambda$ increases the probability of the model sampling a word with the same vowel in the previous end vowel set (improving rhyme consistency), a smaller $\gamma$ reduces the chance of a generated end word being chosen again (increasing rhyming word diversity).

\subsubsection*{Internal Rhyme by Vowel Modeling}
\begin{figure}[ht]
\begin{center}
\centerline{\includegraphics[width=0.8\columnwidth]{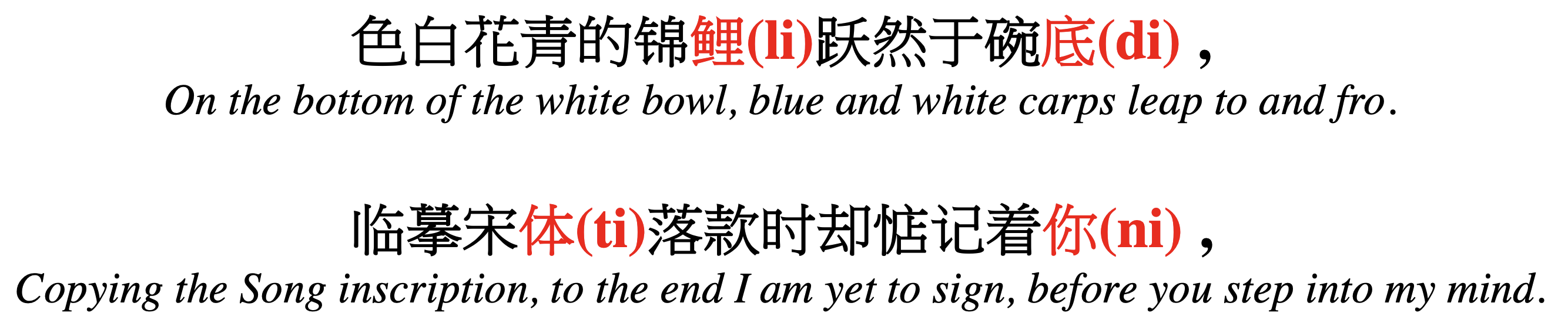}}
\caption{An example of lyrics with internal rhyme. The lyrics are from \textit{Blue and White Porcelain}, a Chinese song of \textit{Jay Chou}, with the English translation underneath. The rhyming characters are highlighted in red with their \textit{pinyin}.}
\label{lyrics-sample-1}
\end{center}
\end{figure}

While rhyme generally refers to end rhyme, where the last words of the lines rhyme with each other, internal rhyme has also been widely used, which usually includes multiple rhyming words either within the same line (usually one in the middle and the other one at the end) or in the middle of multiple lines. As shown in Figure \ref{lyrics-sample-1}, the highlighted characters (with the corresponding \textit{pinyin} in parentheses) share the same vowel and rhyme. While end rhyme has been widely investigated for both poetry generation \citep{lau-etal-2018-deep, li-etal-2020-rigid} and rap lyrics generation \citep{xue-etal-2021-deeprapper, nikolov-etal-2020-rapformer}, internal rhyme is still yet to be explored. 

To model internal rhyme as well as other rhyme schemes with specified vowels, we propose a \textit{restricted vowel loss} to enable direct control of vowels at arbitrary positions. Specifically, as shown in Figure \ref{model}, during the training stage, for the masked fragments in the input, we only replace 80\% of the vowel inputs with the $[M]$ token. For the remaining $20\%$ masked input tokens with ground truth vowel inputs, we introduce an additional vowel prediction task.  For the $t$th output token with a predicted distribution $p_t\in \mathcal{R}^{N\times 1}$, if the ground truth vowel $v_t$ is provided in the input, the additional training objective is

\begin{equation}
   \mathcal{L}^v_t = - \operatorname{log} p(v_t|y_{<t}, \operatorname{enc}(x_{1:T})) \\
\end{equation}

\noindent where the probability  of the vowel $v_t$ at the time step $t$ is calculated by,
\begin{equation}
    p(v_t|\cdot) = \sum_i^N p^i_t \cdot \mathbb{1}(V(x_i)=v_t)
\end{equation}

\noindent where $p^i_t$ is the predicted probability for each token $i$ in the vocabulary at the decoding time step $t$, and $V(x_i)$ is a mapping function which returns the vowel of the token $x_i$. The function $\mathbb{1}(V(x_i)=v_t)$ returns $1$ if the vowel of the token $x_i$ is identical to the ground truth vowel $v_t$, otherwise returns $0$. Thus, $p(v_t|\cdot)$ is basically summing up the predicted probabilities of all the tokens with the same vowel $v_t$.

During the inference stage, the internal rhyme scheme can be created for the generated outputs by providing rhyming vowel inputs to specific positions in masked fragments of the inputs.\footnote{An example is shown in Figure \ref{lyric-sample-partial-rewrite} in Appendix \ref{appendix:examples}.}

\section{Experiment}
\subsection{Datasets and Baselines}
Our model is trained on three different datasets. We first crawl a large-scale text corpus from \textit{Baidu Encyclopedia} , which is the largest Chinese online Encyclopedia. We use this dataset to pretrain our model. Then, we crawl a lyrics dataset from two Chinese lyrics websites.\footnote{\url{https://mojim.com} and \url{https://www.mulanci.org}} 
Since the amount of the lyrics data is limited, we further crawl proses as a supplementary dataset from a Chinese prose website.\footnote{\url{https://www.sanwenwang.com/sanwen/}} 
The pretrained model is first fine-tuned on the prose dataset and then on the the lyrics dataset to produce our final system.\footnote{Details of the datasets are in Table \ref{dataset-table} in Appendix \ref{appendix:dataset}.}

Since Chinese is a monosyllabic language, each character consists of one syllable. To control the number of syllables in the generated output, We use the \textit{BasicTokenizer} from the \textit{transformers} library \citep{huggingface-transformers} for tokenization,  which splits  text into characters for Chinese and into words for other languages (mainly English).  We keep the words and characters with frequency larger than 3,000 to build a vocabulary of size 6,572. For the vowels, we employ \textit{python-pinyin}\footnote{{\url{https://github.com/mozillazg/python-pinyin}}} to extract the vowels of the Chinese characters. There are in total 21 distinct vowels.

We evaluate the proposed model, \textit{SongRewriter}, for both full and partial rewriting tasks. For the full rewriting task, we compare \textit{SongRewriter} with a Chinese \textit{GPT2} \citep{gpt2} and \textit{SongNet} \citep{li-etal-2020-rigid}. For the partial rewriting task, we compare the proposed model with \textit{ILM} \citep{donahue-etal-2020-enabling}. We incorporate the keyword prompt function to the \textit{ILM}  model resulting in \textit{ILM-Keyword} which is used as a baseline model for keyword-conditioned rewriting tasks. All the baseline models are fine-tuned on our lyrics dataset for a fair comparison.\footnote{The details of the baselines are in the Appendix \ref{appendix:baselines}.}

\begin{table*}[ht]
\begin{center}
\scalebox{0.97}{
\small
\begin{tabular}{|c|c|c|c|c|c|c|c|c|c|}
\hline
\multicolumn{1}{|c|}{Model}      &$\Delta$ Diversity $\downarrow$    &$\Delta$ Coherence $\downarrow$    &$\Delta$ PPL-Gen $\downarrow$    &PPL-Test $\downarrow$    &Dist-RW $\uparrow$    &Rhyme-L $\uparrow$    &Rhyme-G $\uparrow$  \\
\hline\hline                                                                    
GPT2                             &0.135                              &0.076                              &3.695                            &3.539                    &0.437                 &0.546                 &0.723               \\
SongNet                          &0.110                              &0.046                              &2.569                            &3.429                    &0.465                 &0.605                 &0.766               \\ 
\hline
SongRewriter                     &\textbf{0.023}                     &\textbf{0.040}                     &\textbf{1.287}                   &\textbf{3.157}           &\textbf{0.552}        &\textbf{0.737}        &\textbf{0.812} \\
\hline
\end{tabular}
}
\end{center}
\caption{Quality evaluation on the generated outputs in the full song rewriting task. The best scores are in bold.}
\label{automatic-full-gen}
\end{table*}

\begin{table*}[ht]
\small
\begin{center}
\begin{tabular}{|c|c|c|c|c|c|c|c|}
\hline
Mask      &Model   & $\Delta$ Diversity $\downarrow$     &$\Delta$ Coherence $\downarrow$      &$\Delta$ PPL-Gen $\downarrow$    &Dist-RW $\uparrow$   &Rhyme-L $\uparrow$   &Rhyme-G $\uparrow$ \\
\hline\hline
\multirow{2}{*}{TOKEN}    & ILM           &0.265                    &0.052               &17.88             &0.547              &0.658              &0.762          \\
                          & SongRewriter  &\textbf{0.039}           &\bf 0.006           &\bf 2.37          &\bf 0.572          &\bf 0.722          &\bf 0.800      \\
\hline
\multirow{2}{*}{SENT}     & ILM           &0.057                    &0.024               &\bf 0.70          &0.500              &0.658              &0.781          \\
                          & SongRewriter  &\bf 0.043                &\textbf{0.002}      &1.66              &\textbf{0.578}     &\textbf{0.742}     &\textbf{0.814} \\
\hline
\end{tabular}
\end{center}
\caption{Quality evaluation on the generated outputs on the task of partial song rewriting under masking schemes, \{SENT, TOKEN\}. We report the averaged scores over three masking ratios, \{0.25, 0.5, 0.75\}. The best scores are in bold.}
\label{partial-rewrite-content}
\end{table*}

\subsection{Evaluation Metrics}
We evaluate the performance in terms of two aspects, generation controllability and generation quality. The controllability metrics include \textit{Keyword Recall} and \textit{Vowel Accuracy}. The quality metrics include \textit{Diversity},\footnote{\textit{Diversity} is evaluated by \textit{distinct} \citep{li-etal-2016-diversity}.}
\textit{Coherence},\footnote{We measure \textit{Coherence} based on sentence similarity.}
Perplexity-Test (\textit{PPL-Test}) and Perplexity-Gen (\textit{PPL-Gen}).\footnote{Details on these metrics can be found in Appendix \ref{appendix:metrics}.} 
Following prior works, we assume the optimal approach should generate lyrics with quality closest to the human-written lyrics \citep{nucleus_sampling}. Therefore, we report the absolute difference scores on some metrics: $\Delta$\textit{Diversity}, $\Delta$\textit{Coherence} and $\Delta$\textit{PPL-Gen}. 

To measure rhyme quality, we propose three novel metrics:

\begin{itemize}
    \item \textbf{Local Rhyme (Rhyme-L)}: While a lyric may contain multiple rhyming vowels, sentences sharing the same rhyme are usually grouped together. Therefore, we propose \textit{local-rhyme-n} to evaluate this localised characteristics. Specifically,  We define a sentence being \textit{locally n-rhymed} if, among the $n$ sentences before and after the current sentence, there are at least one sentence sharing the same rhyming vowel with the current sentence. Thus, \textit{local-rhyme-n} is defined as the number of locally n-rhymed sentences divided by total number of sentences. We report \textit{Rhyme-L}, which is the average of the \textit{local-rhyme-n} with $n\in [1, 4]$.
    
    \item \textbf{Global Rhyme (Rhyme-G)}: Apart from evaluating rhyming effect from a local perspective, we also measure the rhyming effect of a lyrics as a whole. Since the more sentences sharing the same rhyming vowels, the better the rhyming effect. We evaluate the global rhyming performance by calculating the portion of duplicated rhyming vowels: $1 - \frac{\text{number of unique rhyming vowels}}{\text{total number of sentences}}$.
    
    \item \textbf{Diversity of Rhyming Words (Dist-RW)}: Since rhyming with identical words is considered inferior, we evaluate the diversity of the rhyming words by calculating the ratio of the number of unique end words to the total number of end words.  
\end{itemize}

\subsection{Results and Discussion}

\subsubsection*{Full Song Rewriting}
We evaluate the performance of \textit{SongRewriter} on the task of full song rewriting by masking all the input tokens. This task is similar to lyrics generation with a fixed format, i.e., the number of sentences and lengths of each sentence are pre-defined. 

As shown in Table \ref{automatic-full-gen}, \textit{SongRewriter} significantly outperforms other models in terms of both rhyme quality and content quality. For the content, the generated outputs of \textit{SongRewriter} are closer to the human-written lyrics in the aspects of lexical diversity, coherence and fluency (\textit{PPL-Gen}), and \textit{PPL-Test} further verifies the languge modeling ability of \textit{SongRewriter}. 

By inspecting the generated outputs,\footnote{Figure \ref{lyric-sample-full-gen-1} in Appendix \ref{appendix:examples} shows an example.} we observe that: 1) The generated lyrics are fluent and coherent in general. 2) Most lines share the same vowel and rhyme with their neighbouring lines with diverse rhyming words. 3) Similar to human-written lyrics, automatically generated lyrics contain duplicated blocks, which indicates that \textit{SongRewriter} can learn the structural information of the lyrics.

\subsubsection*{Partial Song Rewriting}
We test the performance of \textit{SongRewriter} on the partial song rewriting task by masking a portion of the input lyrics. We compare the proposed model with \textit{ILM} under two masking schemes, \{SENT, TOKEN\}. We average the scores under three masking ratios, \{0.25, 0.5, 0.75\}.\footnote{Detailed results are in Table \ref{partial-rewrite-content-appendix} in Appendix \ref{appendix:partial_rewrite}.}

As shown in Table  \ref{partial-rewrite-content}, the proposed model performs better than \textit{ILM} in general at both token-level and sentence-level masking. Specifically, for \textit{ILM}, it is observed that there is a large discrepancy on content quality (\textit{Diversity}, \textit{Coherence} and \textit{PPL-Gen}) between the TOKEN and SENT masking schemes.On the contrary, \textit{SongRewriter} achieves consistent performance, indicating that the proposed method is more suitable for both tasks. Besides, we also find that the performance of \textit{ILM} decreases significantly as the masking ratio increases, while \textit{SongRewriter} consistently achieves high performance across various masking ratios. This suggests that \textit{SongRewriter} is able to tackle arbitrary portion of the rewriting consistently. By inspecting examples,\footnote{Figures \ref{lyric-sample-partial-rewrite} and \ref{lyric-sample-sentence-rewrite} in Appendix \ref{appendix:examples}.} we observe that, during the partial song rewriting, \textit{SongRewriter} considers not only the bidirectional context but also the rhyming effects with the input sentences.

\subsubsection*{Keyword Control}

\begin{table}[t]
\small
\begin{center}
\begin{tabular}{|c|c|c|}
\hline
Mask          & Model                      & Keyword Recall $\uparrow$  \\
\hline\hline
\multirow{2}{*}{TOKEN}  & ILM-Keyword                       &0.545 \\
                        & SongRewriter                      &\bf 0.890 \\
\hline
\multirow{2}{*}{SENT}   & ILM-Keyword                       &0.451 \\
                        & SongRewriter                      &\bf 0.877 \\
\hline
\multirow{2}{*}{ALL}    & ILM-Keyword                       &0.851 \\
                        & SongRewriter                      &\bf 0.952 \\
\hline
\end{tabular}
\end{center}
\caption{Evaluation results on the task of keyword-conditioned partial song rewriting under masking schemes, \{SENT, TOKEN, ALL\}. We report the averaged scores over three masking ratios, \{0.25, 0.5, 0.75\}. The best scores are in bold.}
\label{keyword-control}
\end{table}

We test the performance of controlled lyrics rewriting under masking schemes {\{SENT, TOKEN\}}. We report the averaged scores over three masking ratios, \{0.25, 0.5, 0.75\}.

To evaluate the ability of generating content with keywords, we first build a keyword database by using \textit{jieba}\footnote{\url{https://github.com/fxsjy/jieba}} to extract keywords from the training set. We evaluate the controllability by sampling keywords from the database.\footnote{We define the sampling probability of keyword $k_i$ to be the number of occurrence of $k_i$ divided by the number of occurrence of all keywords.} As shown in Table \ref{keyword-control}, \textit{SongRewriter} performs significantly better than the baseline model, \textit{ILM-Keyword}, by a large margin.

\subsubsection*{Rhyme Scheme Control}
To evaluate the rhyme scheme controllability, we evaluate the ability of the model to generate tokens with pre-defined vowels at arbitrary positions. We randomly mask 80\% of the vowel inputs from the masked fragments. For the remaining 20\% masked tokens with vowel inputs, we calculate the ratio of output tokens with the same vowel as the input (\textit{vowel accuracy}). We found that the proposed model is able to consistently generate tokens of pre-defined vowels around 98\% of the time under various masking schemes and masking ratios,\footnote{Detailed results are in Table \ref{automatic-eval-result-vowel-appendix} in Appendix \ref{appendix:rhyme_control}.} implying that the model can generate user-defined rhyme schemes by providing rhyming vowel to the target positions in the input most of the times. 

\begin{figure}[t]
\begin{center}
\centerline{\includegraphics[width=\columnwidth]{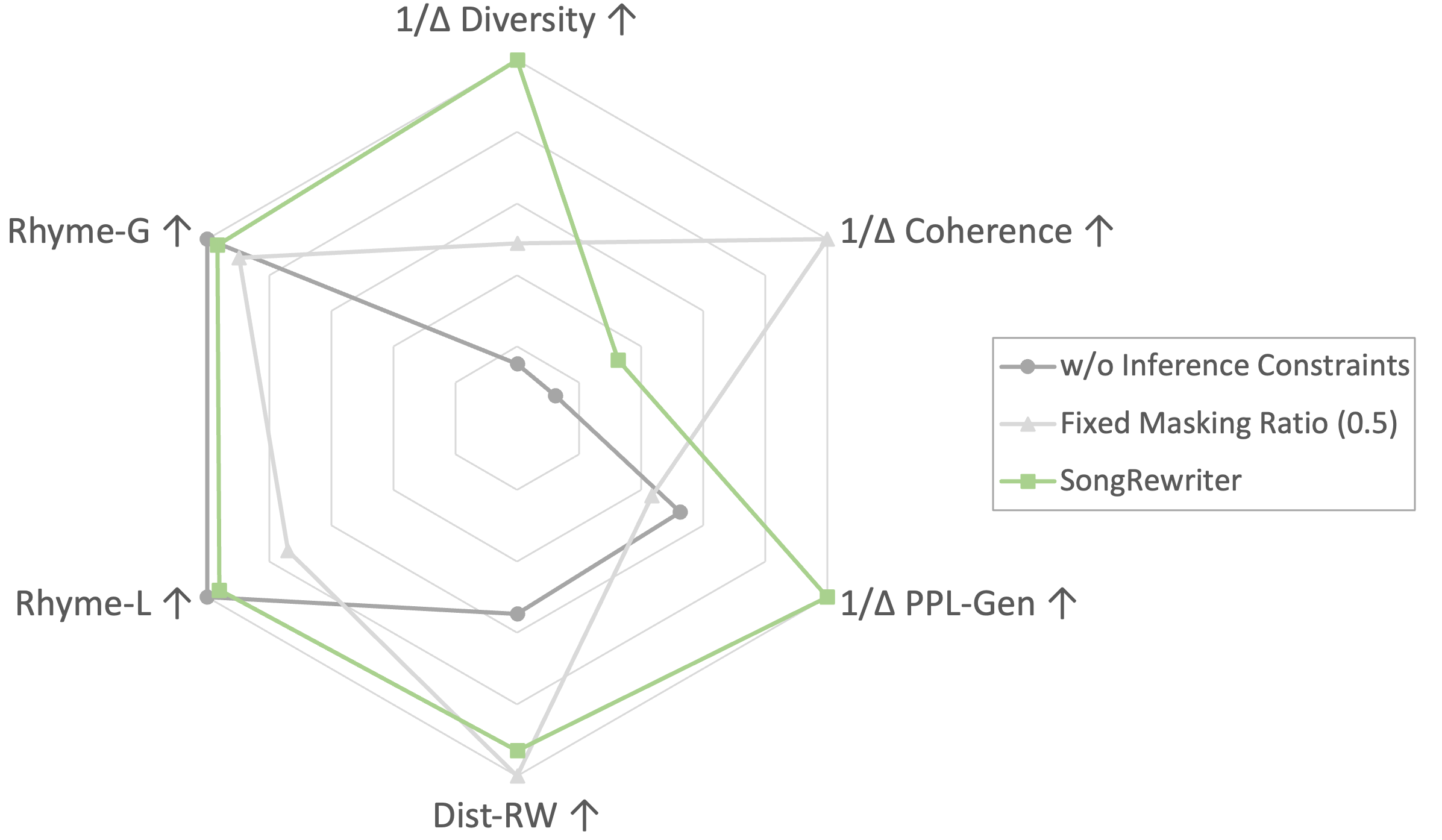}}
\caption{Ablation results on the generated outputs on the task of full song rewriting. To facilitate comparison, all metrics are normalized to 0-1 range by the respective maximum values. The exact scores are presented in Table \ref{ablation-fuill-song} in Appendix \ref{appendix:ablation}.}
\label{ablation-rada}
\end{center}
\end{figure}

\subsubsection*{Ablation Study}
As shown in the Figure \ref{ablation-rada}, by removing the inference constraints, the performance on  \textit{Rhyme-L} and \textit{Rhyme-R} increases while the others decrease. By inspecting the output samples, it is found that the model without inference constraints is more likely to generate repetitive sentences and rhyming words, thus leading to an increase in \textit{Coherence} (more similar content),  \textit{Rhyme-L} and \textit{Rhyme-R}, but a decrease in content diversity and rhyming word diversity. 

To verify the effectiveness of the multi-level masking scheme, we train a model with the masking scheme proposed in MASS \citep{mass}, which is to mask 50\% of the tokens consecutively from the inputs. As shown in Figure \ref{ablation-rada}, the language fluency and the rhyming performance drop significantly in terms of \textit{$\Delta $ PPL-Gen}, \textit{Rhyme-L} and \textit{Rhyme-G}. Although the diversity of the rhyming words increases, those ending words do not share the same vowel, and thus, not rhyme with each other. The performance drop is expected as by only masking 50\% of tokens from the inputs, there is a task discrepancy between the training task (rewriting half of the tokens) and inference task (rewriting content with ratios ranging from 0 to 1). 

Regarding the vowel modeling,  by removing the vowel loss from the training objective, the vowel accuracy for the full song rewriting drops from 98.5\% to 92.5\%, showing that incorporating vowel loss can help the model generate tokens with correct vowels at the specific positions.

\begin{table}[t]
\small
\begin{center}
\scalebox{0.9}{
\begin{tabular}{|c|c|c|c|c|}
\hline
\multicolumn{1}{|c|}{Mask}      &Model                  & Fluency           & Coherence         & Rhyme                \\
\hline\hline
\multirow{2}{*}{TOKEN}          &ILM                    &2.63               &2.89               &2.58                   \\
                                &SongRewriter           &\textbf{3.58}      &\textbf{3.34}      &\textbf{3.75}          \\
\hline
\multirow{2}{*}{SENT}           &ILM                    &3.36               &2.96               &2.91                   \\
                                &SongRewriter           &\textbf{3.71}      &\textbf{3.17}      &\textbf{3.82}          \\
\hline
\multirow{3}{*}{ALL}            &GPT2                    & 3.31              &   3.18            & 2.56              \\
                                &SongNet                 & 3.59             &   3.25            & 3.11              \\
                                &SongRewriter            &\textbf{3.76}     &\textbf{3.38}      & \textbf{3.93}              \\
\hline
\end{tabular}
}
\end{center}
\caption{Human evaluation results on the generated outputs. For the partial song rewriting, the masking ratio is set to $0.5$.}
\label{human-eval}
\end{table}

\subsubsection*{Human Evaluation}
Apart from automatic metrics, we also conduct human evaluation following the previous works \citep{lee-etal-2019-icomposer, xue-etal-2021-deeprapper}. We sample 200 examples from the test set as inputs and generate 200 outputs from each model. Then, we recruit $3$  annotators with musical knowledge background to score the generated lyrics from 1 (Poor) to 5 (Perfect) on three criteria: language  fluency, content coherence and rhyme quality. 

As shown in Table \ref{human-eval}, the human evaluation results show that \textit{SongRewriter} outperforms other models on all three tasks (full song rewriting, sentence rewriting and partial sentence rewriting). In particular, \textit{SongRewriter} performs significantly better in terms of rhyming.

\section{Conclusion}
In this work, we propose to overcome the difficulties of modelling the melody-lyrics correlation from limited parallel datasets by directly rewriting the lyrics of songs. We propose a unified model for full and partial song rewriting by training with a multi-level randomized masking scheme. The proposed model allows rewriting arbitrary parts of the inputs according to the bidirectional context. Besides, we introduce a partial vowel masking strategy into training to enable lyrics generation on any rhyme schemes. A novel decoding strategy is designed to improve the end rhyme consistency and rhyming word diversity. Novel metrics are proposed for rhyme evaluation. Both automatic and human evaluation shows our proposed model outperforms baseline and state-of-the-art models.

\section{Limitation}
Since each Chinese character contains 1 syllable, our proposed model can control the number of syllables in the generation by the number of generated tokens. However, this method does not apply to languages with multisyllabic words (such as English).  To rewrite lyrics with multisyllabic words while maintaining the same number of syllables, a special technique such as syllable-level subword tokenization may be needed. This line of work will be left to be investigated in the future.

\section{Ethics Statement}
Rewriting the lyrics of a song may cause potential copyright infringement. Besides, the copyrights of the lyrics in the dataset belong to the song writers. To protect the copyrights, our model and the released dataset will be protected by the license, Creative Commons Attribution-NonCommercial (CC-BY-NC), and prohibited from commercial use.

\FloatBarrier
\bibliographystyle{acl_natbib}
\bibliography{custom, anthology}

\clearpage
\appendix

\section{Dataset Statistics}
\label{appendix:dataset}

\begin{table}[ht]
\begin{center}
\begin{tabular}{c|c|c|c}
\hline
\multicolumn{1}{|c|}{Model}    &\#train    &\#dev   &\#test     \\
\hline\hline
Baidu Encyclopedia             & 15M       & --     & -- \\
\hline
Prose                          & 330k      & 2k &2k \\
\hline
Lyrics                         & 154k     & 1k & 1k \\
\hline 
\end{tabular}
\end{center}
\caption{Statistics on the number of documents, proses and lyrics. The dev and test sets are randomly sampled.}
\label{dataset-table}
\end{table}

\section{Baselines}
\label{appendix:baselines}
We evaluate the proposed model, \textit{SongRewriter}, for both full and partial rewriting tasks. For the full rewriting task, we compare \textit{SongRewriter} with \textit{GPT2} \cite{gpt2} and \textit{SongNet} \cite{li-etal-2020-rigid} which are fine-tuned on our lyrics datasets. For the partial rewriting task, we compare the proposed model with \textit{ILM} \cite{donahue-etal-2020-enabling}. The details of the baselines are as follows,
\begin{itemize}
    \item \textbf{GPT2}: \textit{GPT2} is an auto-regressive language model based on transformer decoder. Initializing with \textit{GPT2-Chinese},\footnote{\href{https://huggingface.co/uer/gpt2-chinese-cluecorpussmall}{uer/gpt2-chinese-cluecorpussmall}} we fine-tune the model on the lyrics dataset. Note that the lyrics generated by \textit{GPT2} is in free form, it does not follow any format constraints. 
    \item \textbf{SongNet}: \textit{SongNet} is a rigid format controlled text generation model which forces the generated output to follow the exact sentence lengths and sentence numbers of the input. We fine-tune their released pre-trained checkpoint\footnote{\url{https://github.com/lipiji/SongNet}} on the lyrics dataset for full song rewriting. 
    \item \textbf{ILM}: \textit{ILM} is a GPT2-based model specialised on the text infilling task. It randomly replaces part of the text by \textit{"[blank]"} tokens, and appends the masked segments (which are concatenated by \textit{"[answer]"} tokens) to the end of the masked input text. For example, given a source text \textit{"She ate leftover pasta for lunch"}, an \textit{ILM} example will be \textit{"She ate [blank] for [blank] [sep] leftover pasta [answer] lunch [answer]"}. We follow their training strategies and fine-tune \textit{GPT2-Chinese} on the lyrics dataset for partial song rewriting. 
    \item \textbf{ILM-Keyword}: We incorporate the keyword-controlled function to the \textit{ILM} model by adding a keyword prompt to the \textit{ILM} example, leading the input to be in the form, \textit{"[keyword] pasta [keyword] lunch [CLS] She ate [blank] for [blank] [sep] leftover pasta [answer] lunch [answer]"}. We use \textit{ILM-Keyword} as a baseline model for keyword-conditioned tasks. 
\end{itemize}

\section{Training and Inference Settings}
\label{appendix:settings}
The proposed model is a transformer encoder-decoder model \cite{transformer}. There are $12$ layers for the encoder and decoder, respectively, with $12$ heads for each layer. The hidden dimension is $768$, and the dropout \cite{dropout} is $0.1$. We employ the AdamW \cite{adamw} optimizer with a weight decay rate to be $10^{-4}$. For the pre-training stage, we use 8,000 warm-up steps with the default learning rate schedule in  \citet{transformer} and train for $10,000$ iterations.  On the fine-tuning stage, we use a fixed learning rate of $10^{-5}$ and train until models converge. We first fine-tune the model on the prose dataset. Afterwards, we fine-tune the model on the lyrics dataset.  
During inference, we apply \textit{Top-K sampling} \cite{fan-etal-2018-hierarchical} with $k$ to be $32$. The rhyme factors, $\gamma$ and $\lambda$, are set to $0.3$ and $1.4$, respectively.

\begin{table*}[ht]
\begin{center}
\small
\scalebox{0.97}{
\begin{tabular}{|c|c|c|c|c|c|c|c|c|c|}   
\hline
\multicolumn{1}{|c|}{Model}   & Diversity   & Local-STS   & Global-STS   & PPL-Gen   &PPL-Test   & Dist-RW   & Rhyme-L   & Rhyme-G   \\
\hline\hline
Test Set                      & 0.584       &0.276        &0.269         &6.192      &NA         &0.540      &0.724      &0.804       \\
GPT2                          &$0.449_{0.005}$    &$0.343_{0.002}$   &$0.353_{0.001}$   &$2.497_{0.013}$   &$3.539$   &$0.437_{0.005}$    &$0.546_{0.004}$   &$0.723_{0.005}$   \\
SongNet                       &$0.474_{0.002}$    &$0.323_{0.002}$   &$0.313_{0.001}$   &$3.632_{0.011}$   &$3.429$   &$0.465_{0.002}$    &$0.605_{0.002}$   &$0.766_{0.001}$    \\ 
\hline
SongRewriter                  &$0.561_{0.002}$    &$0.322_{0.001}$   &$0.302_{0.001}$   &$4.905_{0.035}$   &$3.157$   &$0.552_{0.002}$    &$0.737_{0.002}$   &$0.812_{0.001}$    \\
\hline  
\end{tabular}
}
\end{center}
\caption{Evaluation on the content quality of the generated lyrics in full song rewriting task. For each model, we apply \textit{Top-K sampling} \cite{fan-etal-2018-hierarchical} for five times and report the mean and standard deviation (subscript). }
\label{automatic-full-gen-overall}
\end{table*}

\begin{table*}[ht]
\small
\begin{center}
\scalebox{0.97}{
\begin{tabular}{|c|c|c|c|c|c|c|c|c|c|}
\hline
Model                         &   Mask                  &Ratio          & Diversity          &Local-STS      & Global-STS    & PPL-Gen   &Dist-RW    & Rhyme-L   &Rhyme-G        \\
\hline\hline
Test Set                      &NA                        &NA            & 0.584         &0.276          &0.269          &6.192      &0.540      &0.724      &0.804          \\
\hline
\multirow{6}{*}{ILM}          & \multirow{3}{*}{TOKEN}   & 0.25         &0.657          &0.271          &0.253          &15.99      &0.602      &0.704      &0.783          \\
                              &                          & 0.5          &0.668          &0.292          &0.270          &22.72      &0.594      &0.654      &0.762          \\
                              &                          & 0.75         &0.692          &0.323          &0.293          &33.49      &0.446      &0.616      &0.740          \\  
\cline{2-10}
                              &  \multirow{3}{*}{SENT}   & 0.25         &0.588          &0.281          &0.269          &7.03       &0.548      &0.686      &0.785          \\
                              &                          & 0.5          &0.536          &0.295          &0.281          &6.71       &0.506      &0.656      &0.782          \\
                              &                          & 0.75         &0.467          &0.333          &0.319          &5.44       &0.446      &0.633      &0.777          \\
\hline
\multirow{6}{*}{SongRewriter}  & \multirow{3}{*}{TOKEN}  & 0.25         &0.632          & 0.266         &0.253          &8.30       &0.590      &0.737      &0.802          \\
                              &                          & 0.5          &0.612          & 0.275         &0.265          &8.22       &0.564      &0.710      &0.796          \\
                              &                          & 0.75         &0.623          & 0.276         &0.264          &9.16       &0.563      &0.719      &0.801          \\ 
\cline{2-10}
                              &  \multirow{3}{*}{SENT}   & 0.25         &0.624          & 0.277         &0.264          &7.38       &0.566      &0.718      &0.800          \\
                              &                          & 0.5          &0.614          & 0.278         &0.265          &7.57       &0.559      &0.743      &0.815          \\
                              &                          & 0.75         &0.644          & 0.280         &0.265          &8.60       &0.608      &0.766      &0.826          \\
\hline
\end{tabular}
}
\end{center}
\caption{Evaluation results on the content quality of the generated outputs on the task of partial song rewriting under masking schemes, \{SENT, TOKEN\},  and masking ratios, \{0.25, 0.5, 0.75\}.}
\label{partial-rewrite-content-appendix}
\end{table*}

\section{Definition of Evaluation Metrics}
\label{appendix:metrics}
\begin{itemize}
    \item \textbf{Keyword Recall}: To evaluate the content control ability by keyword prompt, we calculate the keyword recall rate,  which is the percentage of the keywords appearing in the generated outputs. 
    
    \item \textbf{Vowel Accuracy}: To measure the performance of the vowel control, we calculate the percentage of output tokens with correct vowels as the input.
    
    \item \textbf{Diversity}: We evaluate the diversity of the generated lyrics by \textit{distinct-n} \cite{li-etal-2016-diversity}, which is defined as the number of unique n-grams divided by total number of n-grams. We report the average of the \textit{distinct-n} with \textit{n} from 1 to 4: 
    \begin{equation*}
        \text{Diversity} = \frac{1}{4}\sum_{n=1}^4 \text{distinct-n}
    \end{equation*}
    
    \item \textbf{Coherence}: To evaluate the semantic consistency among all the sentences of the generated lyrics, we measure the semantic textual similarity (STS) between all sentence pairs. Specifically, we employ pre-trained \textit{SimCSE} \cite{gao-etal-2021-simcse} to extract sentence embeddings and calculate the cosine similarity:\footnote{We use \href{https://www.sbert.net/}{sentence-Transformers} library with pre-trained model \href{https://huggingface.co/cyclone/simcse-chinese-roberta-wwm-ext}{cyclone/simcse-chinese-roberta-wwm-ext}.}
    \begin{multline*}
        \text{Global-STS} = \\ \frac{2}{N(N-1)} \sum_{i=1}^{N-1} \sum_{j=i+1}^{N} 
        \text{SimCSE}(s_i, s_j)
    \end{multline*}
    Besides, we also calculate the STS between adjacent sentences for the consistency from a local aspect:
    \begin{equation*}
        \text{Local-STS} = \frac{1}{N-1} \sum_{i=1}^{N-1} \text{SimCSE}(s_i, s_{i+1})
    \end{equation*}
    We define the coherence to be the average: 
    \begin{equation*}
        \text{Coherence} = \frac{1}{2} (\text{Global-STS} + \text{Local-STS})
    \end{equation*}
    
    \item \textbf{PPL-Test}: We evaluate the quality of the model by calculating the model perplexity on the test set. 

    \item \textbf{PPL-Gen}: We evaluate the quality of the generated lyrics with the perplexity from a language model, which is a pre-trained Chinese \textit{GPT2} model\footnote{We use the pre-trained model \href{https://huggingface.co/uer/gpt2-chinese-cluecorpussmall}{uer/gpt2-chinese-cluecorpussmall}.} \cite{gpt2} fine-tuned on the lyrics dataset. 
\end{itemize}

\section{Full Song Rewriting}
\label{appendix:full_rewrite}
\begin{figure}[t]
\begin{center}
\centerline{\includegraphics[width=0.65\columnwidth]{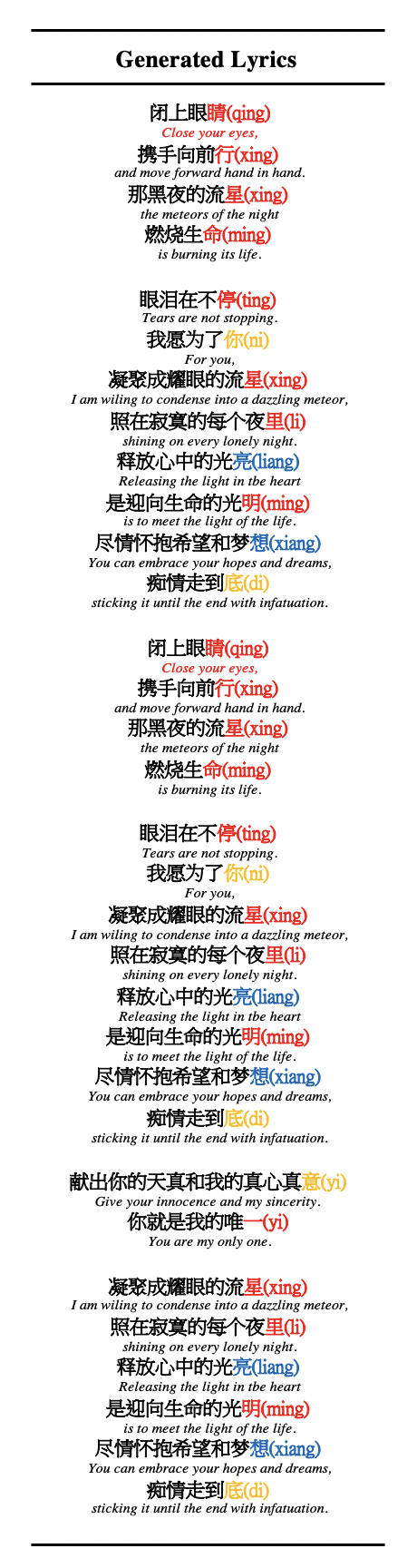}}
\caption{An example generated by \textit{SongRewriter}. End characters sharing the same vowel (the \textit{pinyin} of the characters are in the adjacent bracket) are highlighted in the same color. Lyrics is splitted into blocks for clear illustration. }
\label{lyric-sample-full-gen-1}
\end{center}
\end{figure}

\begin{figure}[t]
\begin{center}
\centerline{\includegraphics[width=0.65\columnwidth]{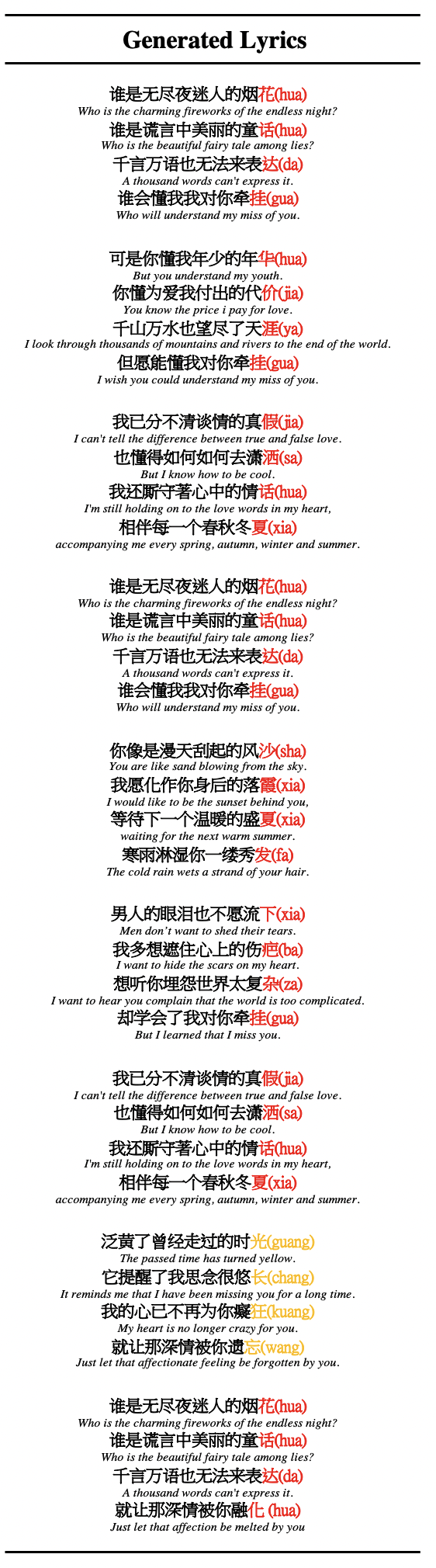}}
\caption{An example generated by \textit{SongRewriter}. End characters sharing the same vowel (the \textit{pinyin} of the characters are in the adjacent bracket) are highlighted in the same color. Lyrics is splitted into blocks for clear illustration. }
\label{lyric-sample-full-gen-2}
\end{center}
\end{figure}

Table \ref{automatic-full-gen-overall} shows evaluation results on the content quality of the generated lyrics in full song rewriting task. For each model, we apply \textit{Top-K sampling} \cite{fan-etal-2018-hierarchical} for five times and report the mean and standard deviation (subscript). Figures \ref{lyric-sample-full-gen-1} and \ref{lyric-sample-full-gen-2} show two examples of the generated lyrics.

\begin{figure*}[ht]
\begin{center}
\centerline{\includegraphics[width=1.5\columnwidth]{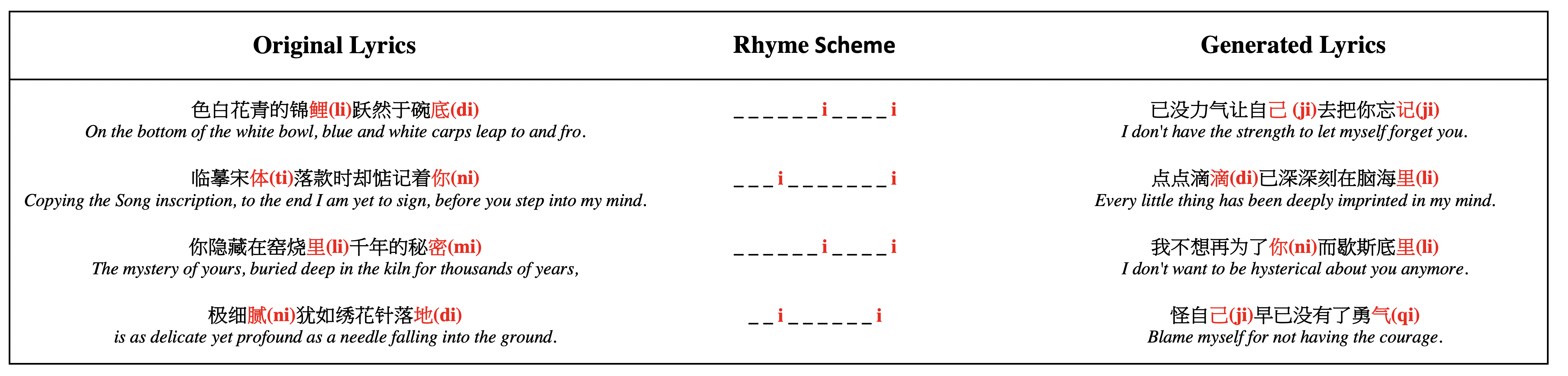}}
\caption{An example of full song rewriting based on the internal rhyme scheme of the input lyrics. The lyrics is from \textit{Blue and
White Porcelain} by \textit{Jay Chou}. The left column is the original lyrics, with the rhyme scheme in the middle column. For the rhyme scheme, only the vowels of the rhyming characters are remained. The vowels of other characters are replaced by a placeholder token. The output lyrics on the right column are generated based on the rhyme scheme.} 
\label{lyric-sample-internal-rhyme}
\end{center}
\end{figure*}

\section{Partial Song Rewriting}
\label{appendix:partial_rewrite}
\begin{figure}[ht]
\begin{center}
\centerline{\includegraphics[width=\columnwidth]{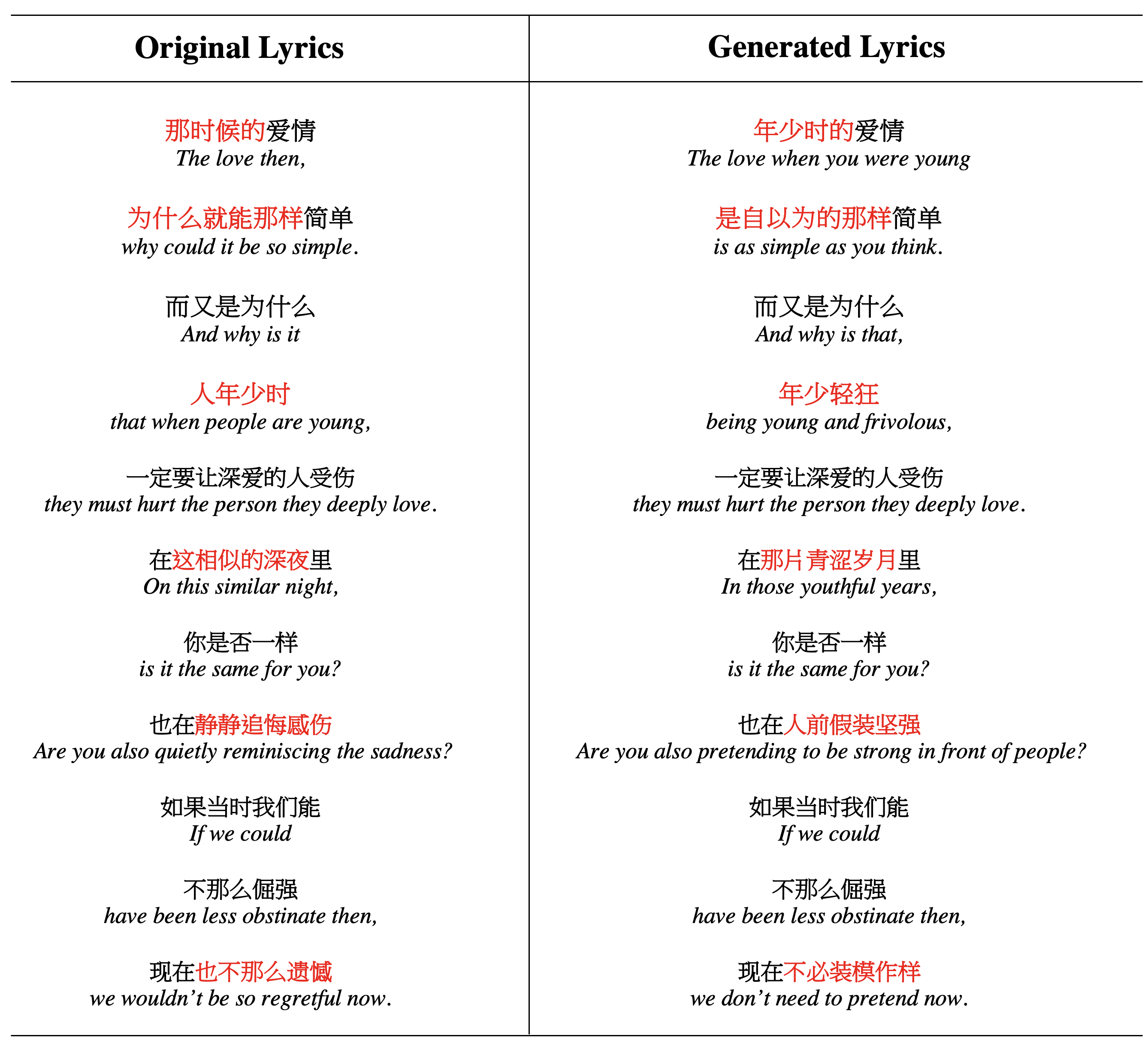}}
\caption{An example of partial sentence rewriting by \textit{SongRewriter}. The original lyrics is from \textit{Later} by \textit{Rene Liu}. The inputs are the original lyrics with red tokens masked. The model generates tokens at the corresponding masked positions (highlighted in red on the right column).  }
\label{lyric-sample-partial-rewrite}
\end{center}
\end{figure}

\begin{figure}[ht]
\begin{center}
\centerline{\includegraphics[width=\columnwidth]{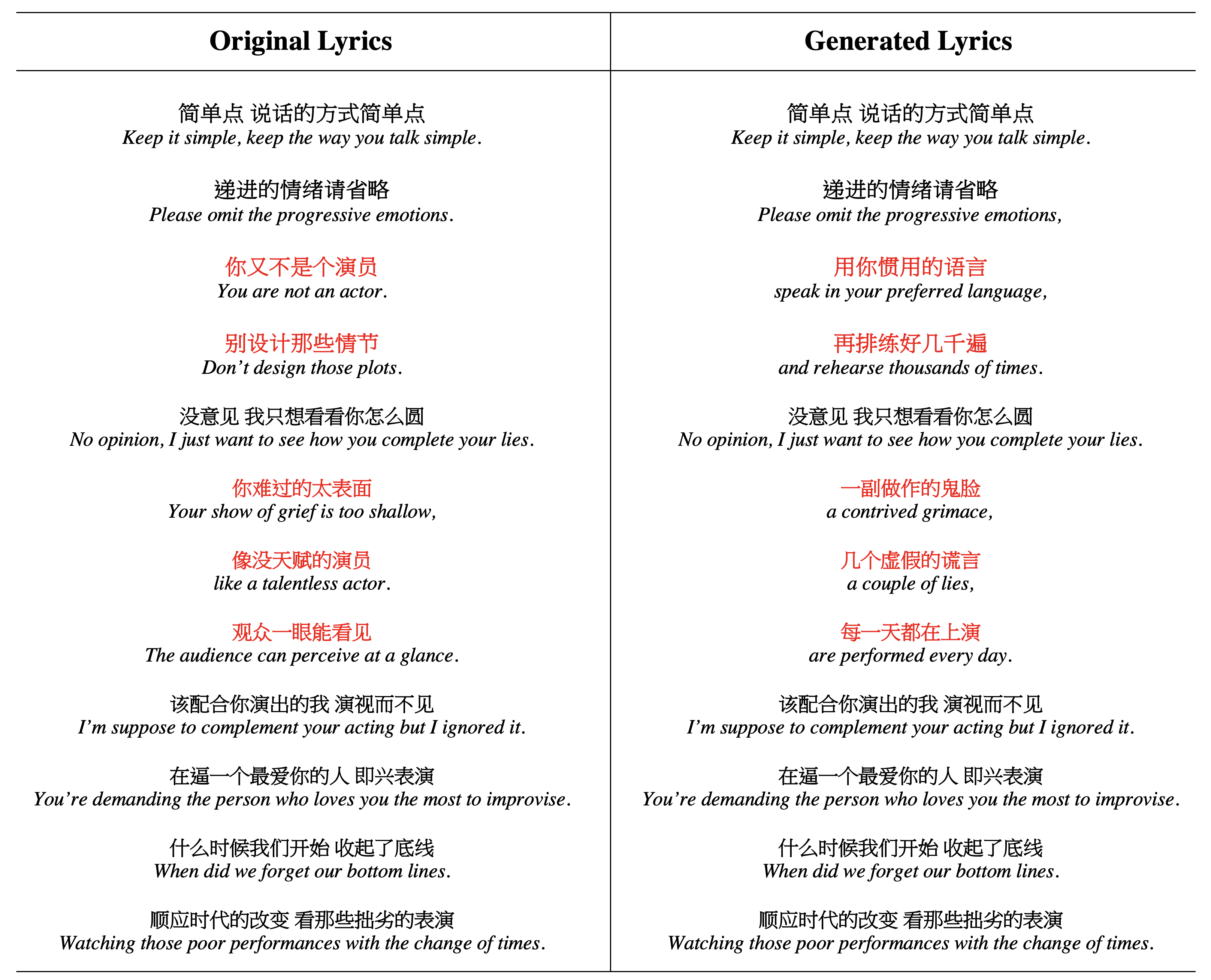}}
\caption{An example of sentences rewriting by \textit{SongRewriter}. The original lyrics is from \textit{Actor} by \textit{Zhiqian Xue}. The inputs are the original lyrics with sentences in red masked. The outputs are the red sentences on the right. }
\label{lyric-sample-sentence-rewrite}
\end{center}
\end{figure}

Table \ref{partial-rewrite-content-appendix} shows evaluation results on the generated outputs on the task of partial song rewriting under masking schemes, \{SENT, TOKEN\},  and masking ratios, \{0.25, 0.5, 0.75\}. Figure \ref{lyric-sample-partial-rewrite} shows an example of token-level rewriting. Figure \ref{lyric-sample-sentence-rewrite} shows an example of sentence-level rewriting.

\begin{table*}[t]
\begin{center}
\small
\scalebox{0.97}{
\begin{tabular}{|l|c|c|c|c|c|c|}
\hline
\multicolumn{1}{|c|}{Model}        &$\Delta$ Diversity $\downarrow$    &$\Delta$ Coherence $\downarrow$      &$\Delta$ PPL-Gen $\downarrow$    & Dist-RW $\uparrow$           & Rhyme-L $\uparrow$       & Rhyme-G $\uparrow$    \\
\hline\hline                      
SongRewriter                       &0.023                         &0.040                            &1.287                 &0.552              &0.737          &0.812 \\
\hline
- inference Constraints            &0.151                            &0.105                              &2.452                   &0.325                       & 0.767                   & 0.839 \\
- Fixed Masking Ratio (0.5)        &0.047                            &0.013                              &2.964                   &0.594                       &0.568                   &0.753  \\
\hline
\end{tabular}}
\end{center}
\caption{Ablation results on the generated lyrics in full song rewriting task. }
\label{ablation-fuill-song}
\end{table*}

\section{Rhyme Scheme Control}
\label{appendix:rhyme_control}
Table \ref{automatic-eval-result-vowel-appendix} shows the evaluation results on the task of vowel-conditioned partial song rewriting under the masking schemes, \{SENT, TOKEN\}, and masking ratios, \{0.25, 0.5, 0.75\}.

Figure \ref{lyric-sample-internal-rhyme} shows an example of applying rhyme scheme control. We first extract the syllable template and rhyme scheme from the original lyrics. As shown in the middle column of Figure \ref{lyric-sample-internal-rhyme}, the original lyrics control both \textit{end rhyme} and \textit{internal rhyme}. The end rhyme scheme is \textit{AAAA}, where all four lines of a verse share the same ending vowel and rhyme with each other. As shown in Figure \ref{lyric-sample-internal-rhyme}, by inputting the desired vowel tokens in the target positions, the generated lyrics have a consistent rhyme scheme with the original lyrics.

\begin{table}[ht]
\small
\begin{center}
\begin{tabular}{|c|c|c|c|}
\hline
Mask          & Ratio             &  Vowel Accuracy  \\
\hline\hline
\multirow{3}{*}{TOKEN}  & 0.25                       &0.984 \\
                        & 0.5                       &0.982 \\
                        & 0.75                       &0.985 \\
\hline
\multirow{3}{*}{SENT}   & 0.25                       &0.984 \\
                        & 0.5                       &0.984 \\
                        & 0.75                       &0.984 \\
\hline
ALL                     & 1.0                       &0.985 \\
\hline
\end{tabular}
\end{center}
\caption{Evaluation results of \textit{SongRewriter} on the task of vowel-conditioned partial song rewriting under masking schemes, \{SENT, TOKEN\}, and masking ratios, \{0.25, 0.5, 0.75\}.}
\label{automatic-eval-result-vowel-appendix}
\end{table}

\section{Ablation Study}
\label{appendix:ablation}
Table \ref{ablation-fuill-song} shows ablation results on the generated lyrics in full song rewriting task.

\begin{table}[ht]
\small
\begin{center}
\begin{tabular}{|l|c|c|c|}
\hline
\multicolumn{1}{|c|}{Model}                         & Dist-RW               & Rhyme-L           & Rhyme-G         \\
\hline\hline
SongRewriter                                        &0.552                  &0.737              &0.812            \\
- Seq TK \& Seq LP                                  &0.500                  &0.510              &0.712              \\
- Seq TK \& Rev LP                                  &0.531                  &0.577              &0.745            \\      
- Rev TK \& Rev LP                                  &0.511                  &0.707              &0.806            \\
\hline  
\end{tabular}
\end{center}
\caption{Evaluation results on the rhyming performance of the generated lyrics on the full song rewriting task under different language order and local position embedding order. }
\label{ablation-language-order-rhyme}
\end{table}

\section{Sequence Order and Local Position Order} 
To improve \textit{end rhyme} modeling, we incorporate \textit{reverse language modeling} with \textit{local position embeddings}. We assume that by reversing the language order, the model is easier to locate the rhyming words (which corresponds to " $\langle l_0 \rangle$", the first token of the reverse sentence according to the local position embedding) and generate rhyming sentences by generating the end words before the rest of the sentences. In this section, we verify the effectiveness of "labeling rhyming words by $\langle l_0 \rangle$" and "generating rhyming word of the sentence first" by comparing four model variants: \textit{reverse token order with sequential local position} (the proposed method),  \textit{reverse token order with reverse local position (Rev TK \& Rev LP)}, \textit{sequential token order with sequential local position (Seq TK \& Seq LP)} and \textit{sequential token order with reverse local position  (Seq TK \& Rev LP)}.

As shown in Table \ref{ablation-language-order-rhyme}, the rhyming performance decreases drastically for models with sequential token order in both \textit{Rhyme-L} and \textit{Rhyme-G}. With the same token order, the models with the $\langle l_0 \rangle$ of the local position aligned with the rhyming word perform slightly better than those not aligned. The results show that generating rhyming words before the other words in the sentences can significantly improve the rhyming performance of the generated outputs. However, the role of local position embeddings to help identify the rhyming words is less important. We hypothesize that, apart from the local position, the model can also identify the rhyming words from the global position (the tokens before or after the sentence delimiter tokens).

\section{Additional Examples}
\label{appendix:examples}

\end{document}